\documentclass[twoside,english,5p]{elsarticle}
\usepackage[T1]{fontenc}
\usepackage{geometry}
\usepackage{amsmath}
\usepackage{amsthm}
\usepackage{stmaryrd}
\usepackage{graphicx}
\usepackage{booktabs}
\usepackage{multirow}
\makeatletter
\theoremstyle{plain}

\theoremstyle{boldremark} % remark bold

\ifx\proof\undefined

\providecommand{\proofname}{Proof}
\fi

\renewenvironment{proof}[1][\proofname]{%
	\par\pushQED{\qed}\normalfont%
	\topsep6\p@\@plus6\p@\relax
	\trivlist\item[\hskip\labelsep\bfseries#1\@addpunct{.}]%
	\ignorespaces
}{%
	\popQED\endtrivlist\@endpefalse
}

\journal{Pattern Recognition Letters}

\usepackage{hyperref}
\hypersetup{colorlinks = true, allcolors = blue}

\usepackage[nameinlink]{cleveref}

\crefname{figure}{Fig.}{Figs.}
\crefformat{equation}{Eq.~#2(#1)#3}
\crefformat{section}{Section~#2#1#3}
\AtBeginDocument{%
\let\citet\cite
}

\usepackage[labelfont=bf]{caption}
\@ifundefined{showcaptionsetup}{}{%
 \PassOptionsToPackage{caption=false}{subfig}}
\usepackage{subfig}
\makeatother

\usepackage{babel}
\providecommand{\remarkname}{Remark}
\providecommand{\theoremname}{Theorem}

\begin{document}

\begin{frontmatter}{}

\title{PETS-SWINF: A regression method that considers images with metadata
	based Neural Network for pawpularity prediction on 2021 Kaggle Competition
	\textquotedbl PetFinder.my\textquotedbl }

\author[rvt]{Yizheng Wang}

\ead{wang-yz19@mails.tinghua.edu.cn}

\author[rvt]{Yinghua Liu\corref{cor1}}

\ead{yhliu@tsinghua.edu.cn}

\cortext[cor1]{Corresponding author}

\address[rvt]{School of Aerospace Engineering, Tsinghua University, Beijing 100084,
	China}

\begin{abstract}
Millions of stray animals suffer on the streets or are euthanized
in shelters every day around the world. In order to better adopt stray
animals, scoring the pawpularity (cuteness) of stray animals is very
important, but evaluating the pawpularity of animals is a very labor-intensive
thing. Consequently, there has been an urgent surge of interest to
develop an algorithm that scores pawpularity of animals. However,
the dataset in Kaggle not only has images, but also metadata describing
images. Most methods basically focus on the most advanced image regression
methods in recent years, but there is no good method to deal with
the metadata of images. To address the above challenges, the paper
proposes an image regression model called PETS-SWINF that considers
metadata of the images. Our results based on a dataset of Kaggle competition,
\textquotedbl PetFinder.my\textquotedbl , show that PETS-SWINF has
an advantage over only based images models. Our results shows that
the RMSE loss of the proposed model on the test dataset is \textbf{17.71876}
but 17.76449 without metadata. The advantage of the proposed method
is that PETS-SWINF can consider both low-order and high-order features
of metadata, and adaptively adjust the weights of the image model
and the metadata model. The performance is promising as our leadboard
score is ranked 18 out of 3537 teams  
for 2021 Kaggle competition on the challenge \textquotedbl PetFinder.my\textquotedbl .
\end{abstract}

\begin{keyword}
	Deep learning\sep Factorization machines \sep Recommender systems
	\sep Fusion \sep Image regression \sep Metadata
\end{keyword}

\end{frontmatter}{}

% 预览段落  14 至  57 段落之源代码

\section{Introduction}

With the development of computers and the application of machine learning,
computer vision is one of the hottest research fields in the data
science world \citet{computer_vision_importance}. Image regression
\citet{image_regression} is different from image classification in
that it requires the model to give a continuous predicted value rather
than a discrete value. There are few researches on image regression,
but in recent years the problem of \textquotedbl facial beauty prediction\textquotedbl{}
(FBP) has been attracting a lot of interest in the pattern recognition
and machining learning communities \citet{FacialBeauty,FacialBeauty1,FacialBeauty2,FacialBeauty3},
and its basic research has contributed to the rapid development of
the plastic surgery and cosmetic industry, such as cosmetic recommendations,
aesthetic surgical planning, facial postural analysis, and facial
beautification \citet{plastic_surgery_and_cosmetic_industry,plastic_surgery_and_cosmetic_industry2}.

But apart from human faces, cute animals also attract people\textquoteright s
attention, which is why more and more people like to keep pets. At
the same time, millions of stray animals suffer on the streets or
are euthanized in shelters every day around the world \citet{Euthanasia1,euthanasial2}.
It would be great if more stray animals can be adopted by caring people
as their pets. So if there is an algorithm that scores images of animals
based on \textquotedblleft facial beauty predictions\textquotedblright ,
we are able to accurately determine a pet photo\textquoteright s attraction
and suggest improvements to give these rescued animals a higher chance
of loving homes.

\begin{figure}
	\begin{centering}
		\includegraphics[scale=0.35]{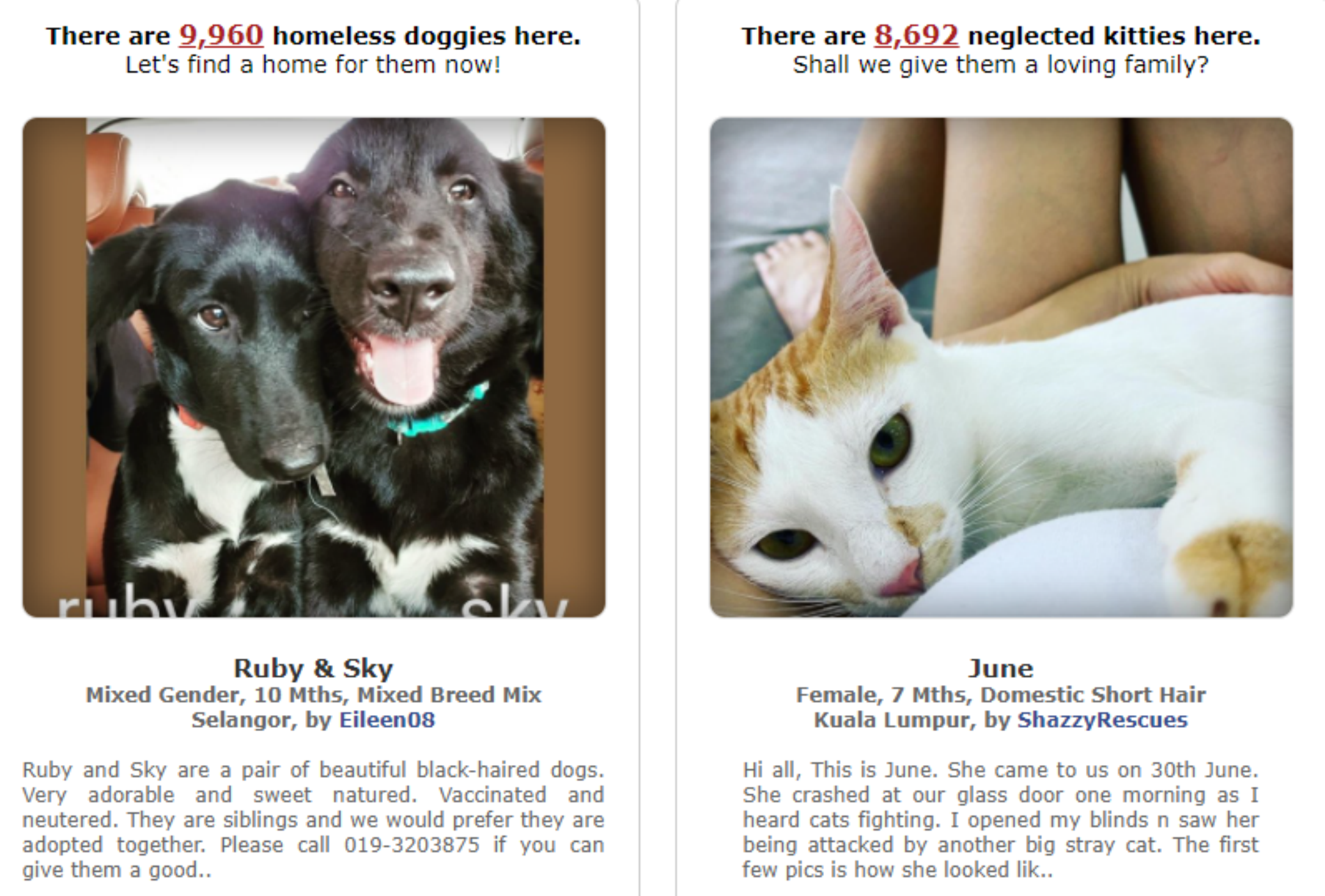}
		\par\end{centering}
	\caption{Petfinder pet profiles (\protect\url{https://www.petfinder.my/}).\label{fig:Petfinder-pet-profiles}}
\end{figure}

The objective of Pawpularity Contest\footnote{https://www.kaggle.com/c/petfinder-pawpularity-score/}
is to help us better understand, analyze, and enhance the pet photos.
This would not only help the animals find loving homes faster, but
also free up shelter and rescuer resources to assist more needy animals.
The situation has become more dire due to the Coronavirus disease,
as shelters are overwhelmed with animals and resources are limited.
In this competition, we analyze raw images and metadata to predict
the \textquotedblleft Pawpularity\textquotedblright{} (cuteness, related
to click-through rate) of stray animals. We train and test our model
on PetFinder.my\textquoteright s\footnote{https://www.petfinder.my/}
thousands of pet profiles, as shown in \Cref{fig:Petfinder-pet-profiles}. 

\textbf{Literature Review}: Image regression differs from image recognition
models as a continuous predictive output is requires, instead of a
discrete one. This category of tasks are rare, where most of them
target at \textquotedbl human facial beauty prediction\textquotedbl{}
such as \citet{plastic_surgery_and_cosmetic_industry2} and \citet{vstvepanek2018evaluation}.
L Xu et al. proposed a new network framework, Classification and Regression
Network (CRNet) \citet{crnet}, which uses different branches to handle
classification and regression tasks simultaneously. Adapting rich
deep features to facial beauty prediction has been applied as feature
extraction techniques reported in \citet{xu2018transferring} and
\citet{lebedeva2021deep}. A residue-in-residual (RIR) structure is
proposed to make gradient flow pass deeper and establish a better
path for information transmission in \textquotedblleft{} Deep Learning
for Facial Beauty Prediction\textquotedblright{} \citet{cao2020deep}.
J Gan et al. propose a lighted deep convolution neural network (LDCNN)
based on the Inception model of Google-net and max-feature-max activation
Layer can extract multi-scale features of images, obtain compressed
representation, and reduce parameters \citet{LDCNN}. Moreover, score-level
fusion methods are proved feasible in recognizing and evaluating animal
facial characteristics \citet{taheri2018animal}. 

Most methods focus on the most advanced image regression methods in
recent years, such as Swin-Transformer \citet{SwinTransformer}, EfficientNet
\citet{efficientnet}. but there is few good methods to deal with
the metadata of images (some abstract features to describe images,
details in \ref{sec:Metadata}). To address the above challenges,
we made two heads for the model, one was predicting pawpularity through
images while the other auxiliary head tried to predict pawpularity
through metadata. Through the performance on the the validation set
to perform the fusion of the two models, we found that the fusion
of the metadata model can improve accuracy of prediction. The main
contributions of this article are two-fold: (i) We propose an image
regression model called PETS-SWINF that considers metadata of the
images. (ii) The model we propose can consider both low-order and
high-order features of metadata, and adaptively adjust the weights
of the image and the metadata model. The proposed method win the in Kaggle competition called 2021 PetFinder.my (currently ranked
\textbf{18/3537}).

The rest of the manuscript is organized as follows: \Cref{sec:Method}
briefly introduces the proposed PETS-SWINF. The data analysis is presented
in \Cref{sec:Data-analysis}. Results and disussion of the proposed
PETS-SWINF are presented in \Cref{sec:Results-and-discussion}. Finally,
\Cref{sec:Conclusion} concludes the work.

\section{Method\label{sec:Method}}

\begin{figure*}
	\begin{centering}
		\includegraphics[scale=0.55]{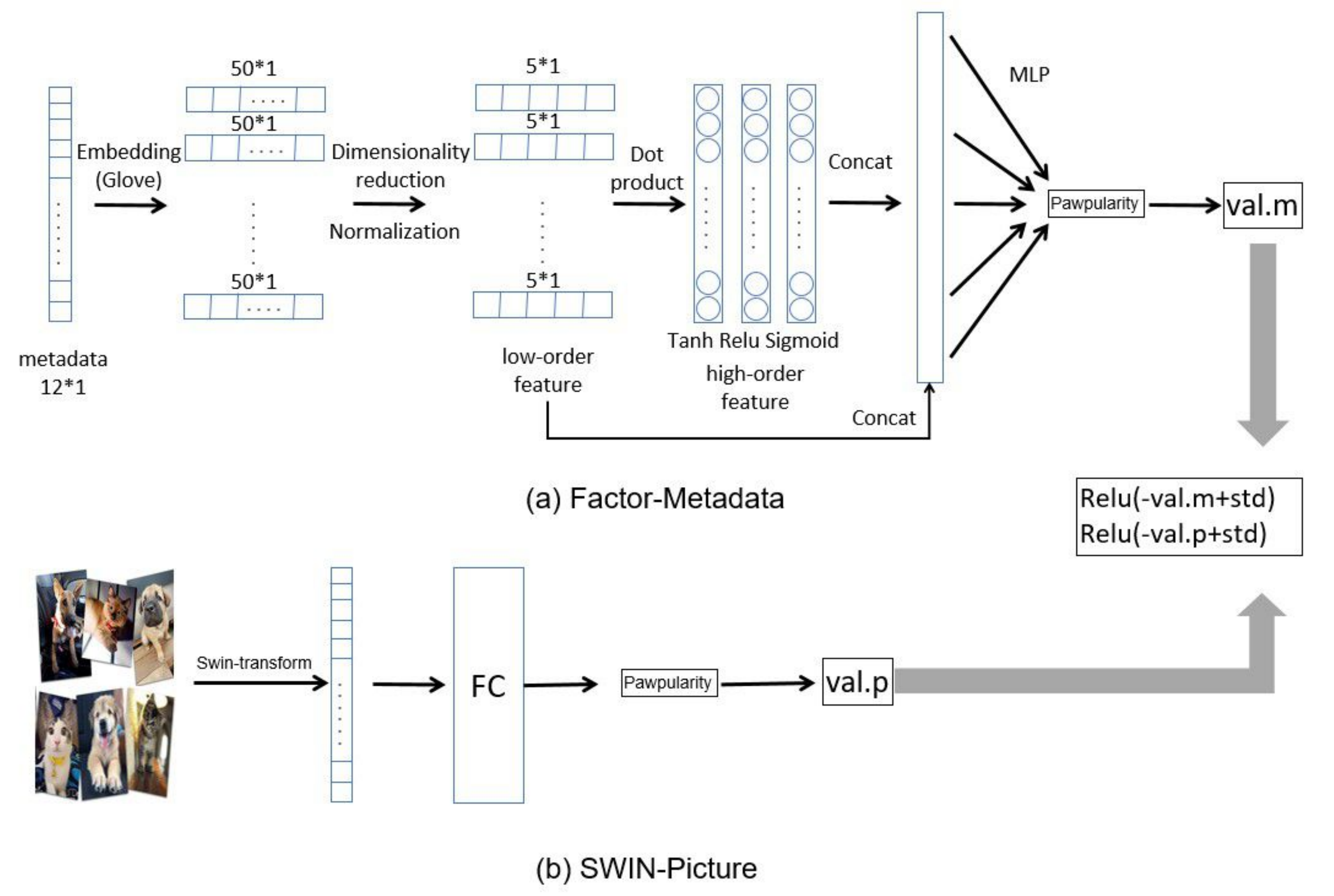}
		\par\end{centering}
	\caption{The proposed PETS-SWINF architecture. (a) Metadata using Factorization-Machine
		(b) image using Swin-Transformer.\label{fig:The-proposed-PETS-SWINF}}
	
\end{figure*}

This problem is essentially a regression problem through images and
metadata. Here we use two single-task learning to regress the images
and metadata separately, and then give the both models weights to
integrate the model. We use Swin Transformer to model the image regression
\citet{SwinTransformer}. Metadata model is inspired by the idea of
DeepFM \citet{DeepFM} for regression. Then the weights of the two
models are obtained based on the loss of the validation set, and the
fusion is carried out. There are three fundamental steps involved
in the proposed regression system- images regression, metadata regression
and parallelly fusion. These steps are shown in \Cref{fig:The-proposed-PETS-SWINF}.

\subsection{Metadata regression \label{subsec:Metadata-regression}}

The flowchart of the metadata model is as shown in \Cref{fig:The-proposed-PETS-SWINF}
(a). We aim to learn both low-order and high-order features of the
metadata. Since metadata have semantic information, we use NLP trained
model embedding (here we use Glove \citet{glove}) instead of processing
the original metadata directly. The reason is that the original metadata
(0 or 1) are not sufficient resulting in the failure to restore the
real input information well if dealing with original metadata. Given
the sparsity of the original data and in order to reduce the latter
computational cost, MLP is used for dimensionality reduction, and
each feature is normalized to obtain low-level features. Next, each
low-level feature is dotted product with all other low-level features,

\begin{equation}
	F_{high}=F_{low}\cdot F_{low}^{T},
\end{equation}
where $F_{low}\in R^{M\vartimes D}$ is the representation of the
lower-order features ($M$ is the number of the metadata, $D$ is
the reduced dimensionality). Then we flatten the high-order features
$F_{high}$, and multiple nonlinear activation functions are applied
to the high-order features. At last the high-order features and low-order
features are concatenated together.

The reason for concatenation is that it\textquoteright s not clear
whether low-level features or high-level features determine the final
regression value. This depends on the target problem. One of the advantages
of this network is that model can learn high-order features and low-order
features based on the characteristics of the dataset. If the two low-order
features are very similiar, then the corresponding high-order feature
is close to 1. The similarity here takes into account the cosine similarity
of the Glove model. The high-order features are acted by the activation
function and can have the ability to the stronger non-linear expression.
After the high-order features and low-order features are concatenated,
MLP is used for regression at last. Finally the RMSE loss of the validation
set is obtained. Here, the loss functions of the validation set and
the training set both use RMSE.

\subsection{Images regression}

Swin-Transformer \citet{SwinTransformer} is adopted as the backbone
network for pawpularity regression. The flowchart of the images model
is as shown in \Cref{fig:The-proposed-PETS-SWINF} (b). We put images
into Swin-Transformer, and then use MLP for regression. After our
experiment, we found that the loss function of BCE is better than
RMSE, so here we use BCE loss as the training loss and RMSE loss as
the validation loss. The BCE loss formula is as follows, 

\begin{equation}
	\mathcal{L}_{BCE}=\frac{1}{N}\sum_{i=1}^{N}[y_{i}\cdot log\tilde{y_{i}}+(1-y_{i})\cdot log(1-\tilde{y_{i}})],
\end{equation}
where $N$ is the number of the data, and $\tilde{y_{i}}$ and $y_{i}$
is the model prediction and the label of the data.

Label of the data is normalized to 0 to 1, although the origin label
of the data is integer 0 to 100. Pretrained model of Swin-transformer
is used before training. The configuration of Swin-Transformer will
be introduced in the \Cref{sec:Results-and-discussion}.

\subsection{Parallelly fusion}

Since two single models are used, we need to assign weights to each
single model. We use the performance of the two models on the validation
set to determine the weight distribution. The weight distribution
adopts the following method,
\begin{equation}
	w_{meta}=\frac{Relu(-val.m+std)}{Relu(-val.m+std)+Relu(-val.p+std)}
\end{equation}
\begin{equation}
	w_{pic}=\frac{Relu(-val.m+std)}{Relu(-val.m+std)+Relu(-val.p+std)},
\end{equation}
where $val.m$ and $val.p$ are the RMSE loss of the validation on
the metadata and the image respectively. $std$ selected as the standard
is the deviation of the training data. The reason why std is used
as a standard value is that if the mean of the label on training dataset
is used as the prediction, then the RMSE will equal to $std$,
\[
std=\sqrt{\frac{1}{N-1}\sum_{i=1}^{N}(label_{i}-u)^{2}}
\]
\[
RMSE=\sqrt{\frac{1}{N}\sum_{i=1}^{N}(label_{i}-pred_{i})^{2}},
\]
where $N$ is the number of the training data, and $u$ is the mean
pawpularity of the training data. Therefore, if we use the mean value
as a prediction, the RMSE loss of this model that does not require
training is actually $std$ when $N$ is enough. So we take $std$
as the worst model performance. If RMSE loss function on the validation
set is smaller than $std$, the more weight is given. If the loss
function is greater than $std$, we set the weight of this model to
zero, because this model will bring negative effects.

\section{Data analysis \label{sec:Data-analysis}}

\begin{figure}
	\begin{centering}
		\includegraphics[scale=0.45]{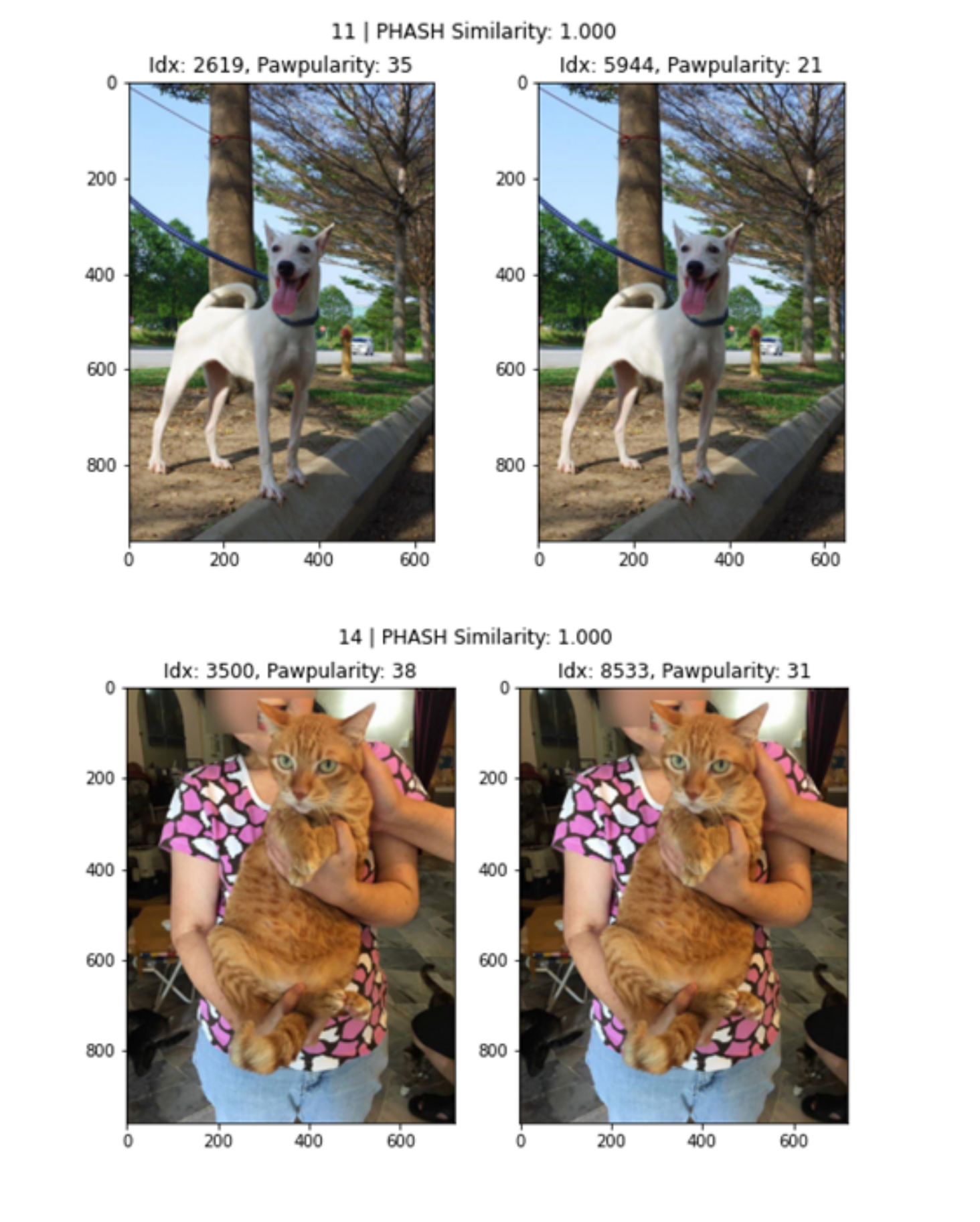}
		\par\end{centering}
	\caption{Similar images: the label of the duplicate images is different.\label{fig:Similar-picture:-the}}
\end{figure}

There are 9912 images with metadata in the training set. The details
of metadata will be introduced in \ref{sec:Metadata}. We find
there are duplicate images in the training set, and the label of the
duplicate image are different. We find the duplicate image by image
perceptual hashing (perceptual hashes are \textquotedbl close\textquotedbl{}
to one another if the images are similar). Because the number of the
duplicate image is 54 as shown in \Cref{fig:Similar-picture:-the}
(the total number of the images is 9912, a small proportion of the
total image), we keep the duplicate images, in order to better make
the model learn the noise of the data. Our experiments show that keeping
duplicate images performs better on the test set than deleting them
\footnote{Data analysis is inspired by https://www.kaggle.com/markwijkhuizen/petfinder-eda-yolov5-obj-detection-tfrecords}. 

\begin{figure}
	\begin{centering}
		\includegraphics[scale=0.25]{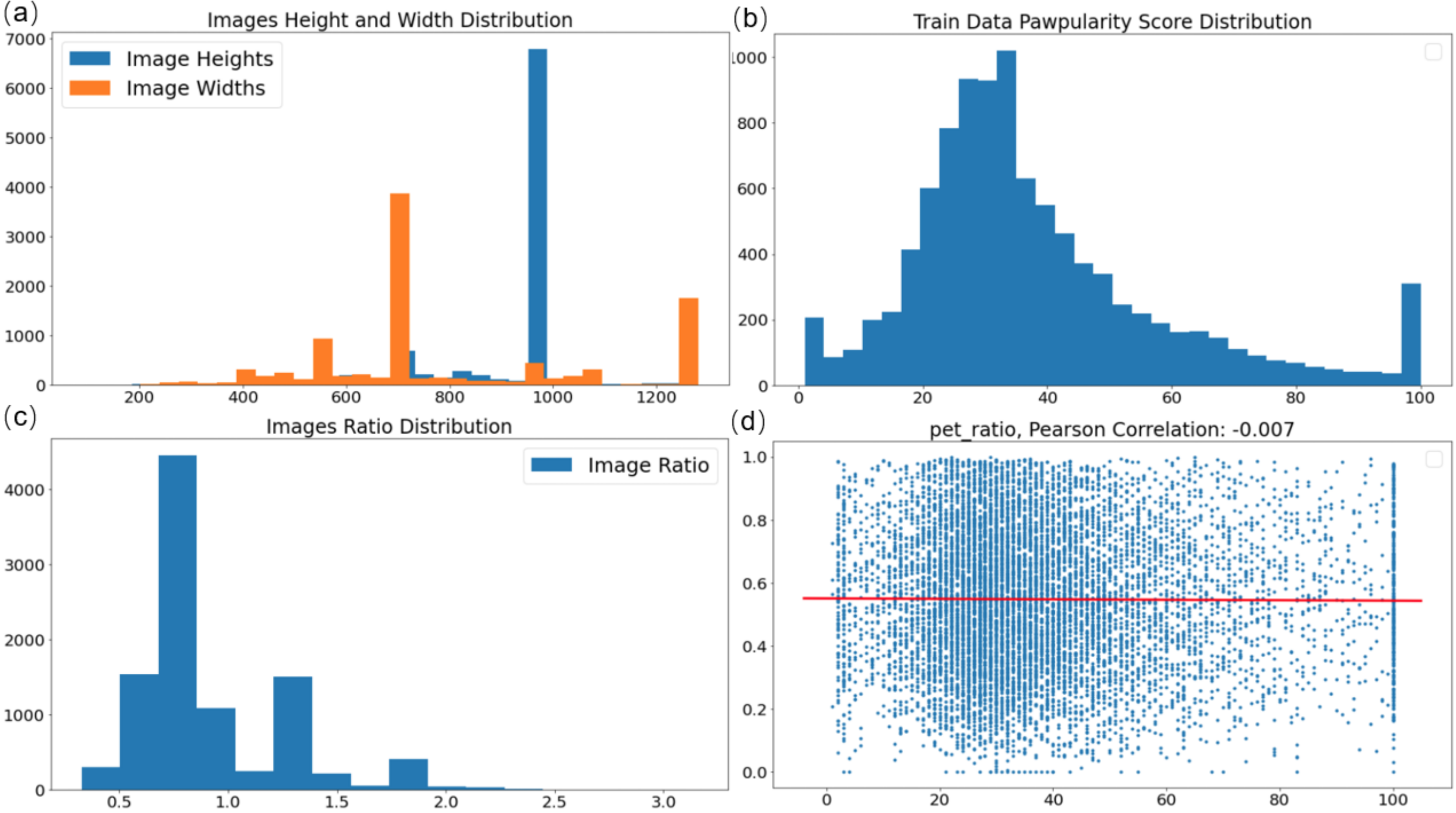}
		\par\end{centering}
	\caption{The statistical characteristic of the data by Exploratory Data Analysis.
		\label{fig:The-statistical-characteristic}}
\end{figure}

The statistical characteristics of the images are analyzed. We analyzed
image height and width distribution, ratio distribution ($width/height$),
and pawpularity distribution on the train data. \Cref{fig:The-statistical-characteristic}
(a) and (c) show the size distribution of the images, we can find
that most of the images are taken by mobile phones. \Cref{fig:The-statistical-characteristic}
(b) shows that the distribution of the pawpularity is similar to the
gauss distribution, and the amount of the data pawpularity at both
ends is particularly large due to the upper and lower limit. Experiments
show that keeping extreme images (pawpularity 100 or 0) is better
than deleting them. \Cref{fig:The-statistical-characteristic} (d)
shows there is a few correlations between the image ratio and pawpularity. 

\begin{figure}
	\begin{centering}
		\includegraphics[scale=0.35]{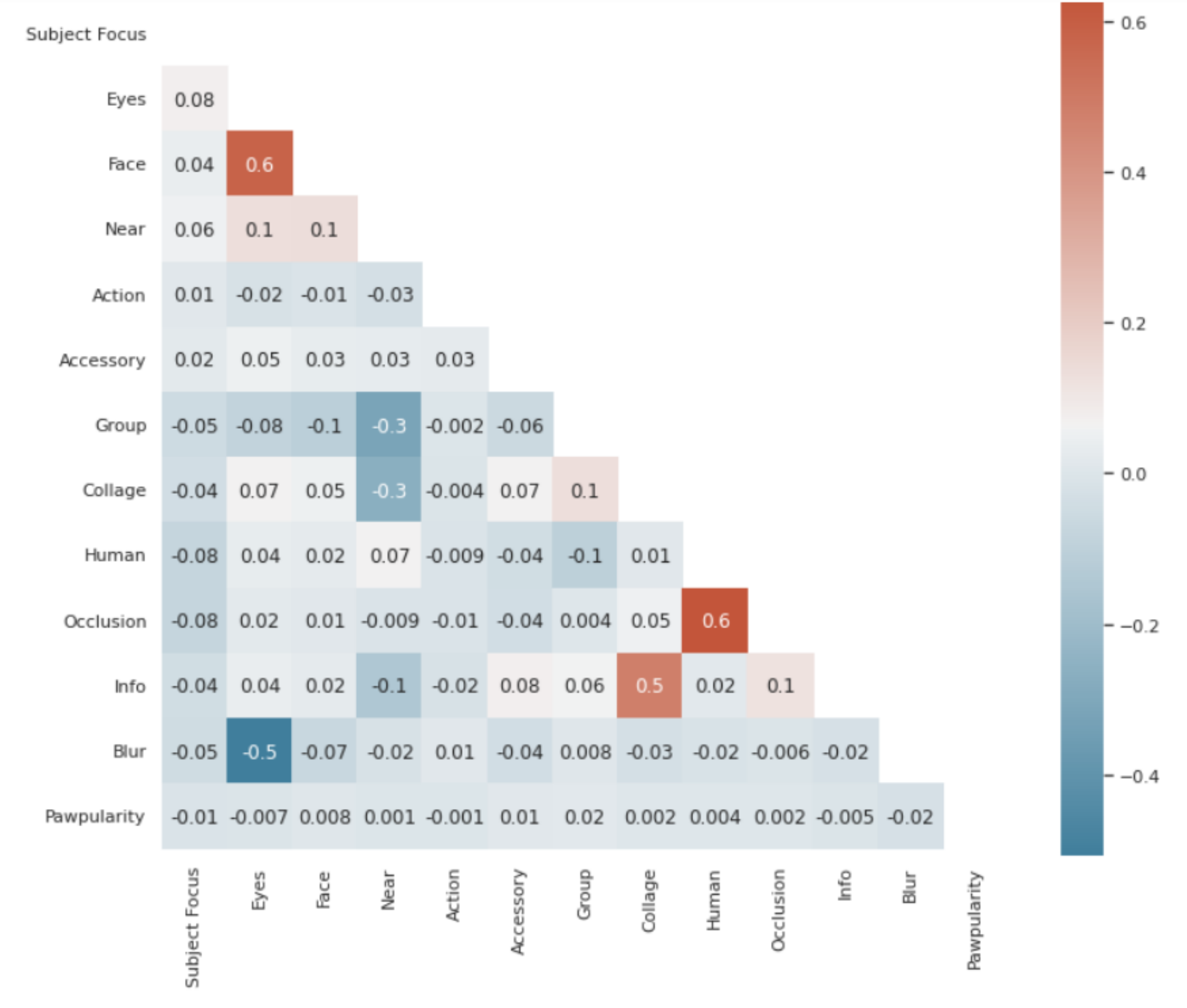}
		\par\end{centering}
	\caption{The correlation between the meta 12 features and the pawpularity.
		\label{fig:The-correlation-between}}
\end{figure}

There are 12 features in the metadata detailed in \ref{sec:Metadata}.
We use Correlation Matrix between the meta features and the pawpularity
as shown in \Cref{fig:The-correlation-between}. We can find that there
are a few relations between all the meta-features and particularity.
Although the linear correlation between metadata and pawpularity is
low, there may exist a nonlinear relationship. Therefore, it is important
to design nonlinearities in metadata networks. This is the reason
for the design of high-order features.

\begin{figure}
	\begin{centering}
		\includegraphics[scale=0.2]{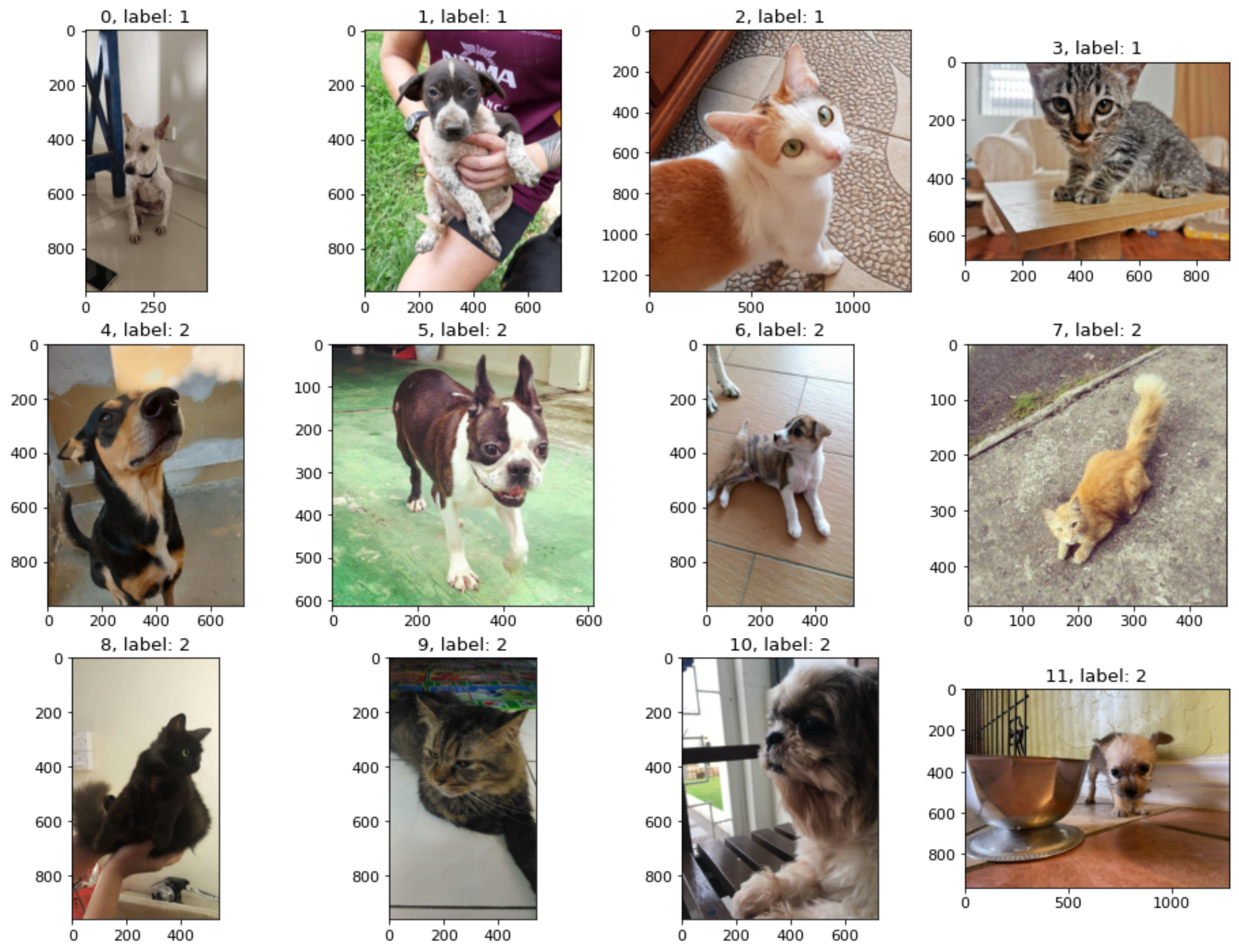}
		\par\end{centering}
	\caption{Lowest pawpularity. \label{fig:Lowest-pawpulartiy.}}
\end{figure}

\begin{figure}
	\begin{centering}
		\includegraphics[scale=0.2]{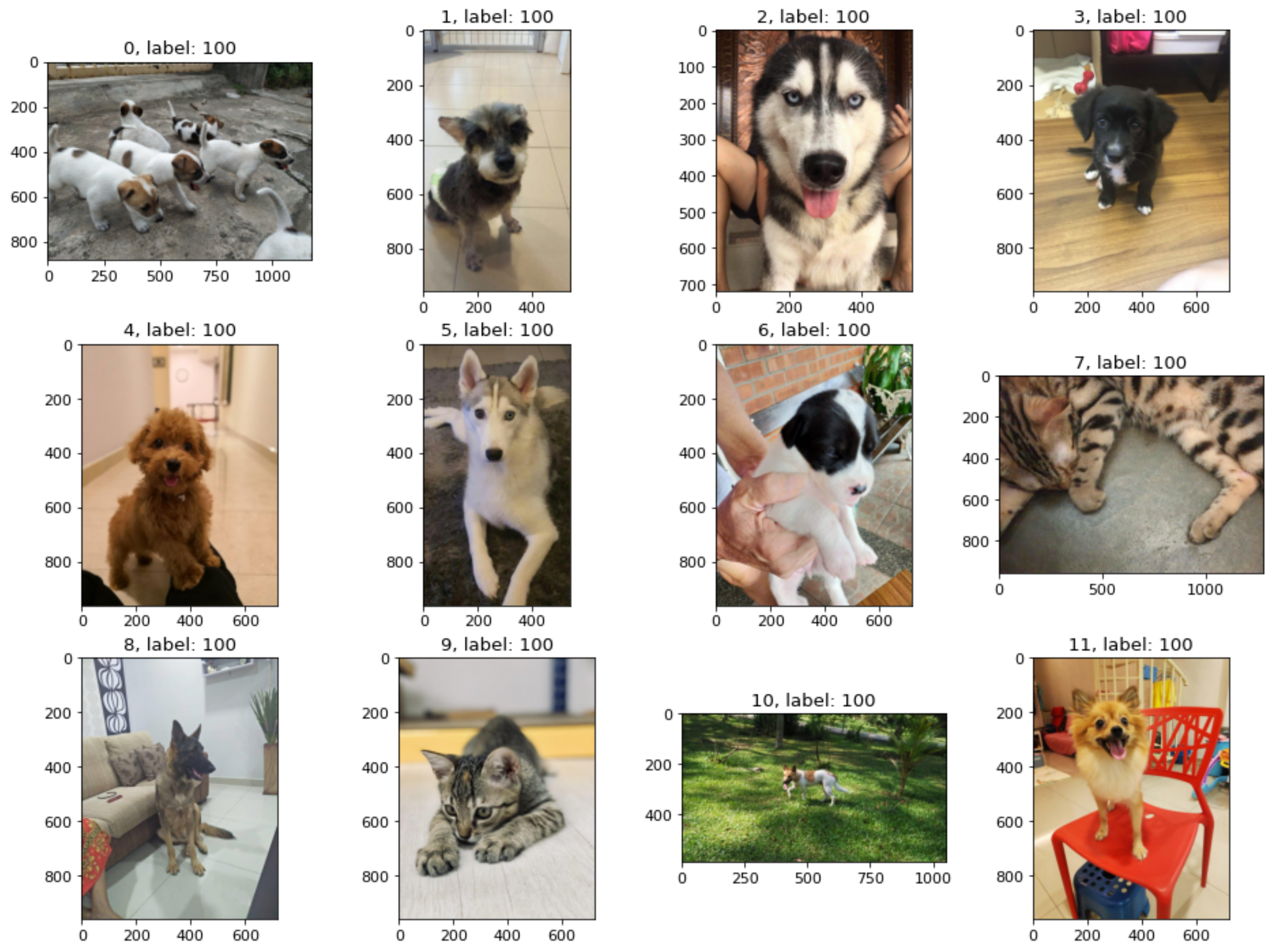}
		\par\end{centering}
	\caption{Largest pawpularity. \label{fig:Largest-pawpulartiy.}}
\end{figure}

The pawpularity label is derived from each pet profile\textquoteright s
page view statistics at the listing pages, using an algorithm that
normalizes the traffic data across different pages, platforms (web
\& mobile) and various metrics. We visualize the extreme cases as
shown in \Cref{fig:Lowest-pawpulartiy.} and \Cref{fig:Largest-pawpulartiy.}.
We can find that there is a big difference between the pawpularity
and our cognitive loveliness (the image 2 in \Cref{fig:Lowest-pawpulartiy.}
should have more pawpularity than its label and the image 8 in \Cref{fig:Largest-pawpulartiy.}
is too average to deserve a full score of 100). The pawpularity score
is related to the time the animal started to be shown in web, because
the derivation of the pawpularity is strongly related to the traffic. 

\begin{figure}
	\begin{centering}
		\includegraphics[scale=0.45]{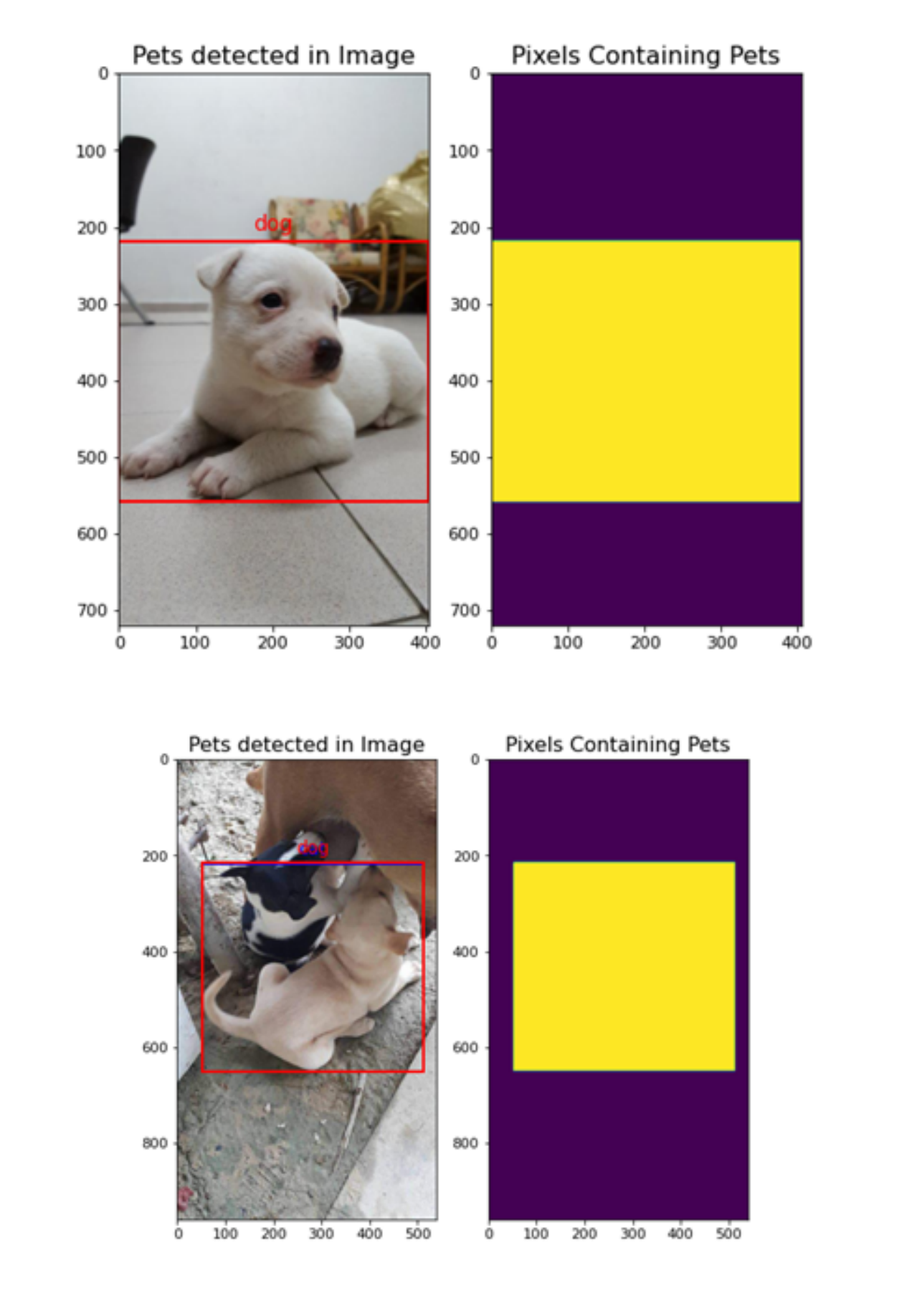}
		\par\end{centering}
	\caption{The result about cats and dogs of the yolov5. \label{fig:The-result-of}}
\end{figure}

\begin{figure}
	\begin{centering}
		\includegraphics[scale=0.2]{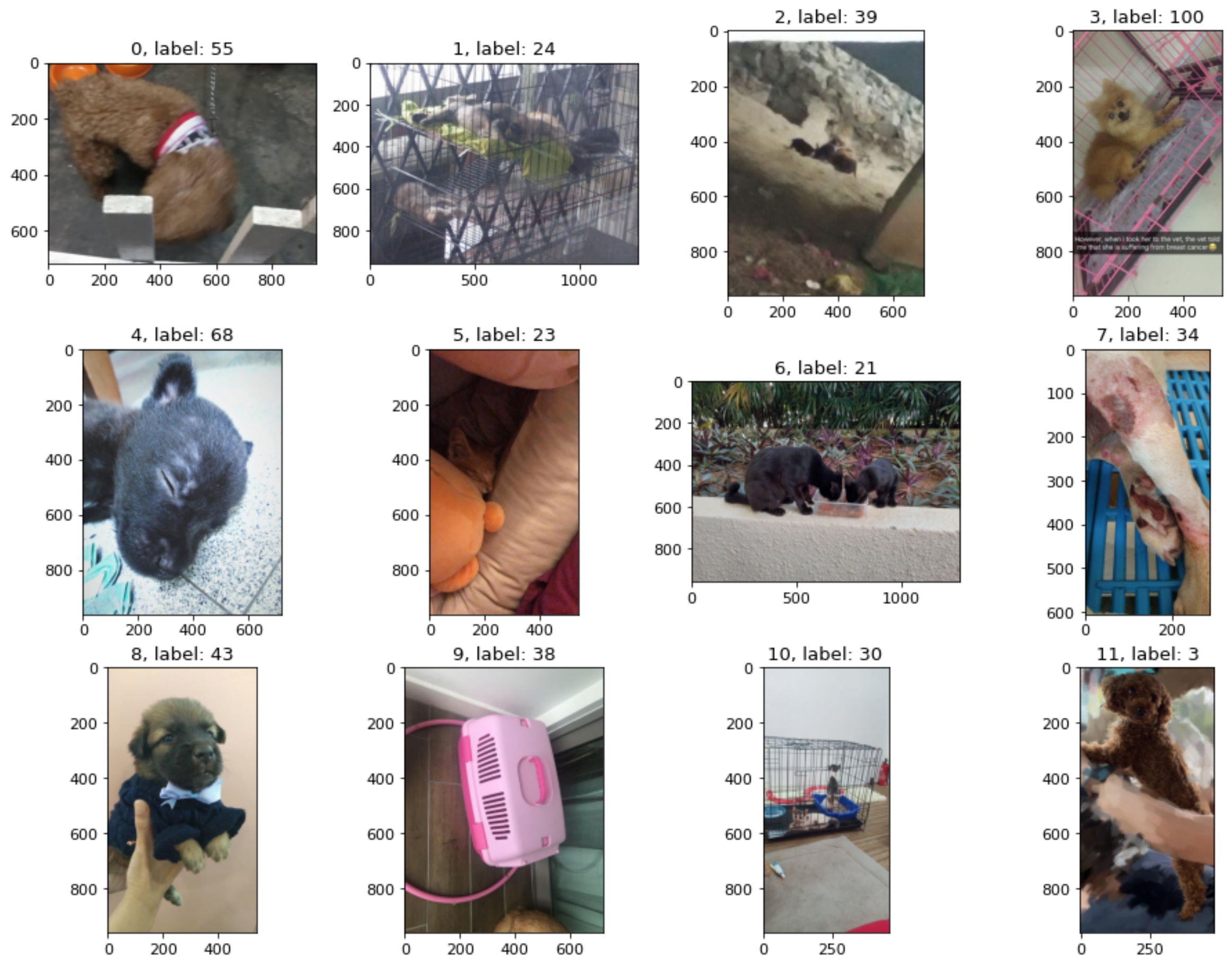}
		\par\end{centering}
	\caption{The result unknown of the yolov5. \label{fig:The-result-of-1}}
\end{figure}

\begin{figure}
	\begin{centering}
		\includegraphics[scale=0.22]{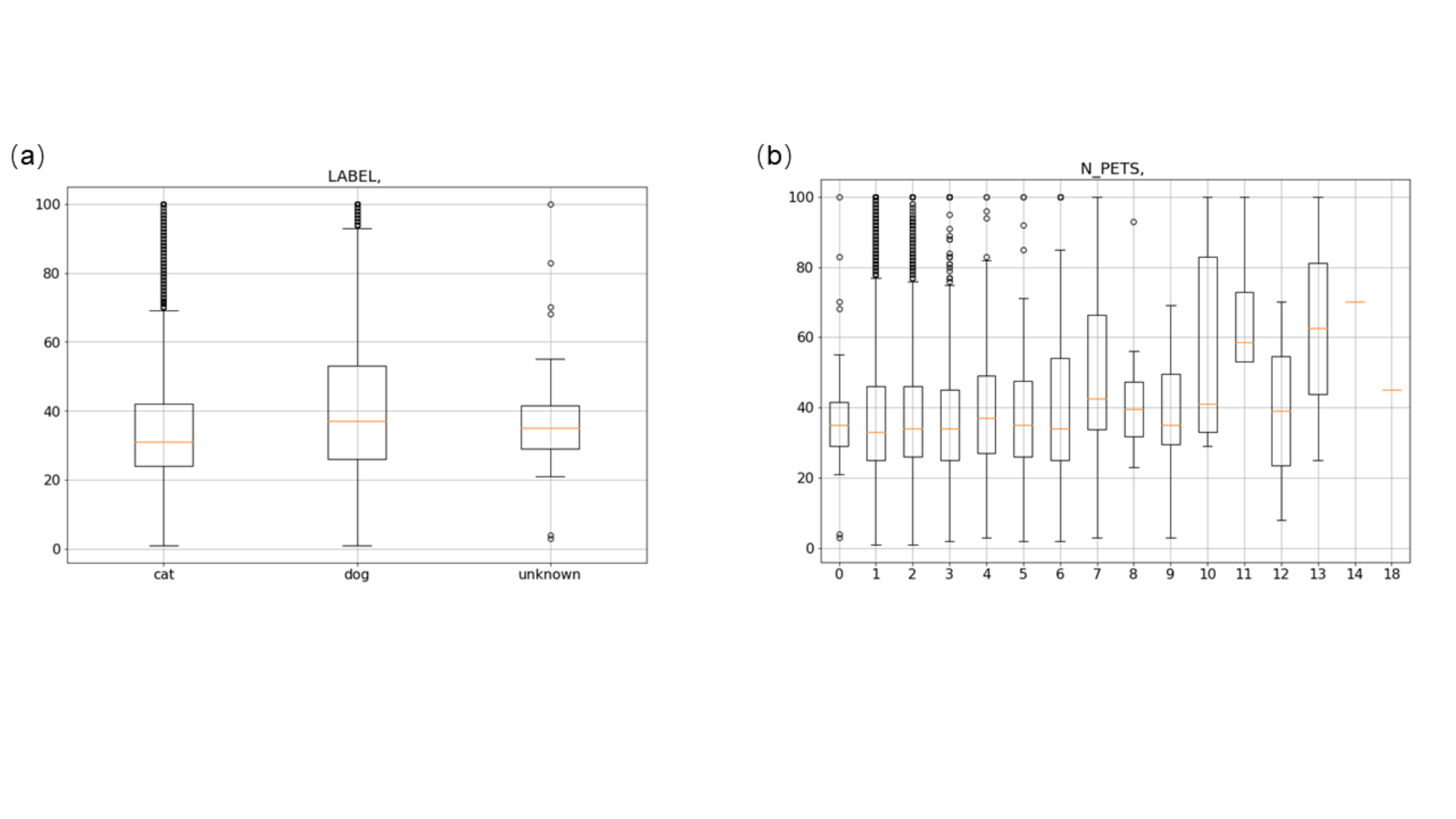}
		\par\end{centering}
	\caption{The correlation between the classification (dogs and cats) and the
		pawpularity (a), the correlation between the number of animals (dogs
		and cats) in one image and the pawpularity (b).\label{fig:The-correlation-between-1}}
\end{figure}

There are only two kinds of animals (dog and cat) in the dataset.
It is reasonable that dogs and cats are judged differently. There
is no classified information in the dataset, so we use the yolov5
(yolov5x6) in object detection to classify the images. \Cref{fig:The-result-of}
shows the results of the yolov5 only about cats and dogs and \Cref{fig:The-result-of-1}
shows the unknown results of the yolov5. \Cref{fig:The-correlation-between-1}
shows that the boxplot of classifications of the image and the number
of the animals in one image both are different. 

\section{Results and discussion \label{sec:Results-and-discussion}}

To conduct our experiments, the proposed PETS-SWINF architecture achieved
RMSE 17.71876 with data augmentations on the test dataset, and win
the gold medal (currently ranked 12/3393). As shown in \Cref{tab:Results-obtained-from},
we compare our results without data augmentations and without metadata.

\begin{table*}
	\caption{Results obtained from PETS-SWINF, PETS-SWINF without data augmentation,
		and PETS-SWINF without metadata. PETS-SWINF without data augmentaions
		has metadata regression models. PETS-SWINF without metadata has data
		augmentation. PETS-SWINF has data augmentation and metadata regression
		models.\label{tab:Results-obtained-from}}
	
	\centering{}%
	\begin{tabular}{lll}
		\toprule 
		& Val accuracy & Test accuracy\tabularnewline
		\midrule
		PETS-SWINF without data augmentaions & 17.68121 & 17.88243\tabularnewline
		PETS-SWINF without metadata & 17.48691 & 17.76449\tabularnewline
		PETS-SWINF & \textbf{17.42164} & \textbf{17.71876}\tabularnewline
		\bottomrule
	\end{tabular}
\end{table*}

We use RandomErasing, Rotate, Brightness, Flip, Contrast, Saturation
as the data augmentation configuration. Batch size is 40; KFLOD is
10; Learning rate is 2e-5. Swin\_large\_patch4\_window7\_224 is adopted
as the backbone network for the image regression. TTA (Test-Time Data
Augmentation ) as the training data augmentation configuration is
used in the inference period.

It can be found from \Cref{tab:Results-obtained-from} that data augmentations
and the rational use of metadata improve the accuracy of the validation
set and the test set. This is because there are not enough data in
the training set, so the data augmentations can have a good effect.
In addition, these augmentation techniques have a certain intersection
with the nature of the pawpularity prediction problem. Although it
can be seen from \Cref{fig:The-correlation-between} that independent
metadata have little effect on pawpularity, it cannot be ruled out
that some nonlinear combinations of metadata are independent to pawpularity.
Therefore, The metadata processing of PETS-SWINF establishes the correlation
between nonlinear features (low-order features and high-order features)
through the neural network and the cuteness. The model uses the loss
of the validation set to perform the weighting fusion, which shows
that the weighting method of model fusion is useful from the result
of PETS-SWINF without metadata and PETS-SWINF.

\section{Conclusion \label{sec:Conclusion}}

Millions of stray animals around the world suffer in the streets or
are euthanized in shelters every day. If more stray animals can be
adopted by caring people as their own pets, it will greatly reduce
the number of stray animals. In order to better adopt stray animals,
it is nessary to score the pawpularity of stray animals, but evaluating
the pawpularity of animals is a very labor-intensive thing. In this
study, we proposed a PETS-SWINF architecture, for prediction of the
pawpularity from pets images with metadata. The proposed method win
the gold medal in Kaggle competition called 2021 PetFinder.my (currently
ranked 18/3537 ).

The proposed method is a regression model, which can combines images
with metadata. The way of assigning weight can adaptively use images
or metadata. In addition, the metadata model can fuse low-order features
and high-order features. The proposed PETS-SWINF model is available
publicly for open access at \url{https://github.com/yizheng-wang/PETS-SWINF}. 

The proposed method can be widely used in image regression tasks with
metadata, such as CTR (click-through rate prediction). As more and
more stray animals are being labeled all around the world, larger
datasets are being generated. We will continue to further modify the
architecture of the proposed PETS-SWINF and incorporate new available
datasets. New versions of the PETS-SWINF will be released upon development
through the aforementioned link.

\section*{Acknowledgements}

The proposed method was inspired by Professor Zhu and Professor Tang
in machine learning class 2021. The study was supported by the Major
Project of the National Natural Science Foundation of China (12090030).
The authors would like to thank Chenxing Li and Yanxi Zhang for helpful
discussions.

\appendix

\section{Metadata\label{sec:Metadata}}
\begin{itemize}
	\item Focus - Pet stands out against uncluttered background, not too close
	/ far. 
	\item Eyes - Both eyes are facing front or near-front, with at least 1 eye
	/ pupil decently clear. 
	\item Face - Decently clear face, facing front or near-front. \textbackslash item
	Near - Single pet taking up significant portion of photo (roughly
	over 50\% of photo width or height). 
	\item Action - Pet in the middle of an action (e.g., jumping).
	\item Accessory - Accompanying physical or digital accessory / prop (i.e.
	toy, digital sticker), excluding collar and leash.
	\item Group - More than 1 pet in the photo. 
	\item Collage - Digitally-retouched photo (i.e. with digital photo frame,
	combination of multiple photos). 
	\item Human - Human in the photo. \textbackslash item Occlusion - Specific
	undesirable objects blocking part of the pet (i.e. human, cage or
	fence). Note that not all blocking objects are considered occlusion. 
	\item Info - Custom-added text or labels (i.e. pet name, description). 
	\item Blur - Noticeably out of focus or noisy, especially for the pet\textquoteright s
	eyes and face. For Blur entries, \textquotedblleft Eyes\textquotedblright{}
	column is always set to 0.
\end{itemize}

\bibliographystyle{elsarticle-num}
\addcontentsline{toc}{section}{\refname}\bibliography{finalreport}

\end{document}